\pdfoutput=1

\documentclass[11pt]{article}

\usepackage[final]{ACL2023}

\usepackage{times}
\usepackage{latexsym}
\usepackage{multirow}
\usepackage{graphicx}
\usepackage{enumitem}
\usepackage{array}
\usepackage{multirow}
\usepackage{booktabs}
\usepackage{graphicx}
\usepackage{adjustbox}
\usepackage{makecell} 

\definecolor{myred}{RGB}{252, 86, 3}
\definecolor{myblue}{RGB}{3, 161, 252}

\usepackage[T1]{fontenc}

\usepackage[utf8]{inputenc}

\usepackage{microtype}

\usepackage{inconsolata}

\title{Customizing Large Language Model Generation Style using Parameter-Efficient Finetuning}

\author{Xinyue Liu \and Harshita Diddee \and Daphne Ippolito \\
  Carnegie Mellon University \\
  \texttt{\href{mailto:xinyuel4@andrew.cmu.edu}{xinyuel4@andrew.cmu.edu}}; 
  \texttt{\href{mailto:harshitadd@gmail.com}{harshitadd@gmail.com}}; 
  \texttt{\href{mailto:daphnei@cmu.edu}{daphnei@cmu.edu}}}

\begin{document}
\maketitle

\begin{abstract}
One-size-fits-all large language models (LLMs) are increasingly being used to help people with their writing. However, the style these models are trained to write in may not suit all users or use cases. LLMs would be more useful as writing assistants if their idiolect could be customized to match each user. In this paper, we explore whether parameter-efficient finetuning (PEFT) with Low-Rank Adaptation can effectively guide the style of LLM generations. We use this method to customize LLaMA-2 to ten different authors and show that the generated text has lexical, syntactic, and surface alignment with the target author but struggles with content memorization. Our findings highlight the potential of PEFT to support efficient, user-level customization of LLMs. 

\end{abstract}

\section{Introduction}



Language models, especially ones trained to be ``human-aligned'' and conversational, are increasingly being used to help people write, including for student essays \citep{bavsic2023chatgpt}, screenplays \cite{mirowski2023co}, stories \citep{ippolito2022creative}, and science communication \citep{bedington2024writing}.
Nearly all of these applications rely on ChatGPT, Gemini, or other LLM instances trained by large companies and shared across all users.
Concerns have been raised that this reliance on a handful of LLMs is leading to a homogenization of language \citep{samual2023what}, hindering students from developing their own writing styles \citep{hasanein2023drivers} and introducing cultural and linguistic biases that fail to reflect diverse backgrounds \citep{ray2023chatgpt}.

Thus, in this paper, we are interested in how language model generations can be customized to the writing styles of individual users.
Our focus is on writers who already have some 1k-50k tokens of prior work (which, in an education setting, could be the writing of an author they are learning to emulate).
Our method aims to create customized LLMs that adopt the idiolect of the target writer while retaining the ability to understand and follow natural language instructions. In addition, users should have the choice of whether customization includes learning ``content'' words such as named entities that are present in the source data.

In the past, when LLMs were smaller, it was common to control the style of generations via finetuning on data within the target style, as \citet{sawicki2022training} do with two Romantic poets, and \citet{van2021fine} do with NPC dialogue.
However, full model finetuning is untenable for today's state-of-the-art language models.
More recently, model customization has been performed via prompt engineering---prefixing a user's query to the model with a set of instructions or exemplars of the target style that is intended to guide the model's outputs~\citep{NEURIPS2020_1457c0d6}.
The success of this technique heavily relies on the prompt's structure~\citep{min2022rethinking} and whether the model's training data contains similar instructions.
Also, an author's prior work may be too large to fit into most LLMs' maximum context lengths.
In contrast to prior approaches, we explore whether a model's generation style can be altered via small amounts of finetuning, using parameter-efficient finetuning (PEFT) methods such as LoRA \citep{hu2021lora}.  
PEFT is a promising direction for model style customization because it eliminates finicky prompt engineering and is efficient to use.

We introduce \textit{StyleTunedLM}, a novel approach that leverages LoRA for efficient finetuning of LLMs to generate text in specific writing styles.
We compare \textit{StyleTunedLM} with prompt engineering and few-shot learning approaches, showing it is more effective at capturing the style of training data.
We also tackle two challenges with tuning on unstructured data---preserving instruction-following ability after finetuning and learning style signifiers without learning content words.

\section{Methods}
\paragraph{StyleTunedLM} 




We build our method by finetuning LoRA adapters for the pre-trained Llama-2-7b model~\citep{Touvron2023Llama2O} on unstructured text datasets from specific authors, using a next-token prediction objective.
The goal is to tailor the model's output to reflect specific stylistic characteristics while maintaining the capabilities learned in prior training.
Finetuning details can be found in Appendix~\ref{app:lora_details}. For style-following generation examples, see \href{https://cauchy221.github.io/Research-StyleTunedLM-Demo/}{https://cauchy221.github.io/Research-StyleTunedLM-Demo/}.

\paragraph{Baselines} 
In our \textbf{fewshot} baseline, we prompt Llama-2-7b with 5 or 10 randomly selected 256-token excerpts from the target author before asking it to generate continuation given a prompt.
In our \textbf{instruct} baseline, we use Llama-2-7b-chat, a variant of the Llama-2-7b finetuned to be conversational~\citep{Wei2021FinetunedLM}.
We prompt with the target author's name and an instruction to generate a continuation in the writing style.
We post-process model outputs to remove irrelevant phrases like ``Please tell me if you have further questions.''

\paragraph{Masking out Named Entities}
Users of customized LLMs ought to be able to control the extent to which their custom model learns words associated with content, rather than style.
In our work, we examine whether certain classes of words, such as names, can be excluded from the learning process.
We first use spaCy~\citep{spacy} to annotate each token position with whether it corresponds to a person's name.
During finetuning, we set the \verb|attention_mask| to $1$ while changing their labels to $-100$ in the loss calculation.
This method could be applied to any class of words that a user prefers the model not to learn.

\paragraph{Merging LoRA Modules}



Building on recent advancements in enhancing pre-trained models with instruction-following capabilities, we propose a novel approach to integrate both style-following and instruction-following functionalities within a single model. This innovation is motivated by the challenge that StyleTunedLMs face in handling tasks requiring a broader understanding of user instructions, such as generating stories with specific elements. We address this by concatenating the weight matrices A vertically and B horizontally, effectively preserving both functionalities. Specifically, we merge a LoRA module fine-tuned on the LIMA instruction dataset~\citep{Zhou2023LIMALI} with a StyleTunedLM. To the best of our knowledge, this is the first approach to enable a fine-tuned model's instruction-following ability by merging LoRA modules.



\section{Experimental Design}
\label{sec:eval-framework}

\paragraph{Author Dataset} Ideally, we would evaluate corpora from authors not present in the training data, as this best reflects the target users of customized models.
However, since most LLMs do not disclose their pre-training data, we conduct an imperfect evaluation using the works of ten authors from Project Gutenberg~\citep{Gerlach2018ASP}.
\ref{sec:app-auth} provides a detailed introduction to each author.
We collect all available books from each author and randomly
divide them into training, validation, and test datasets. The training and validation sets are used for model finetuning and selection, while the test set is reserved for generative tasks used in our evaluation.
Notably, a book assigned to one dataset does not appear in the others.

\paragraph{Evaluation Dataset}
We evaluate in-style generation on a dataset of 100 prompts.
50 prompts were generated using GPT-4, as detailed in~\ref{sec:gpt4}. The remaining 50 prompts were randomly selected from the test set. For each author, we extracted five sentences and used the first 6-8 words of each sentence to create a prompt.

\subsection{Evaluating Generation Style}
\label{sec:framework}

Inspired by earlier studies of author style~\citep{Syed2019AdaptingLM, Verma2019ALS}, we evaluate our stylized generation across three dimensions: perplexity on withheld text, style-embedding alignment, and linguistic alignment. For each prompt in the evaluation dataset, the model is asked to generate a continuation of $256$ tokens.

\paragraph{Perplexity} The capacity of LLMs to understand and generate text consistent with a target author can be measured by the perplexity of withheld text.
We compare the PPL of \textit{StyleTunedLLM}s against the pre-trained Llama-2-7b model across validation sets for each author.


\paragraph{Style-embedding Alignment} 

\begin{table*}[t]
\centering
\small
\begin{tabular}{c|c|cc|cccccc}
\toprule
\textbf{Author} & \textbf{Method} & \textbf{\% in training} & \textbf{\# of names} & \textbf{PPL\(\downarrow\)} & \makecell{\textbf{Cosine} \\ \textbf{Sim.}} & \textbf{Acc.} & \makecell{\textbf{Lexical} \\ \textbf{\scriptsize (MSE)\(\downarrow\)}} & \makecell{\textbf{Syntactic} \\ \textbf{\scriptsize (JSD)\(\downarrow\)}} & \makecell{\textbf{Surface} \\ \textbf{\scriptsize (MSE)\(\downarrow\)}} \\ \midrule

\multirow{2}{*}{PGW} & w/o masking & 0.50 & 68 & 9.68 & 1.0 & 1.0 & 0.18 & 0.07 & 0.01 \\ 
& w/ masking & 0.23 & 91 & 10.46 & 0.98 & 0.9 & 0.16 & 0.07 & 0.11 \\ 
\multirow{2}{*}{JA} & w/o masking & 0.61 & 62 & 7.93 & 1.0 & 1.0 & 7.72 & 0.04 & 12.53 \\ 
& w/ masking & 0.45 & 85 & 8.02 & 0.9 & 0.76 & 4.62 & 0.03 & 7.49 \\ 
\bottomrule
\end{tabular}
\caption{Model performance with and without masking during training of PGW (P. G. Wodehouse) and JA (Jane Austen). With masking, the number of names matching the training data decreases, even as the number of unique names in the generation increases. Masking has minimal effect on style alignment.}
\label{tab:masking}
\end{table*}

Building on prior research in authorship attribution and verification~\citep{Wegmann2022SameAO, Tyo2021SiameseBF}, we train a Sentence-Transformer \citep{reimers-2019-sentence-bert} to embed text excerpts from the 10 authors on our training set. We also use 256 as the sequence length here.
For each author, we compute the average embedding of the text excerpts for the author.
We assess stylistic similarity by measuring the distance between each average embedding and model outputs.
In our preliminary experiment, we compared pairwise and triplet loss for training the style attribution model and chose the former as it led to more separated author clusters (see \ref{sec:all_tsne})

We also finetune a BERT classifier (bert-base-uncased) to classify text excerpts as one of the ten authors.
Together, these dual methods provide a comprehensive validation of the model's style alignment.

\paragraph{Linguistic Alignment} Following the framework of~\citet{Verma2019ALS}, we evaluate our method across three linguistic levels: \textit{lexical}, \textit{syntactic}, and \textit{surface}. Lexical assesses word choice, syntactic reviews sentence structure complexity, and surface examines text's statistical features, with details in Appendix~\ref{app:linguistic_details}. 
For measuring style alignment, we use Mean Squared Error (MSE) for lexical and surface levels and Jensen-Shannon Divergence (JSD) for syntactic analysis, which provides a probability distribution vector. These metrics collectively quantify the unique stylistic features of an author's writing style.




\section{Experiment Results}


\paragraph{Perplexity}

\begin{figure}[t]
\centering
\includegraphics[width=0.4\textwidth]{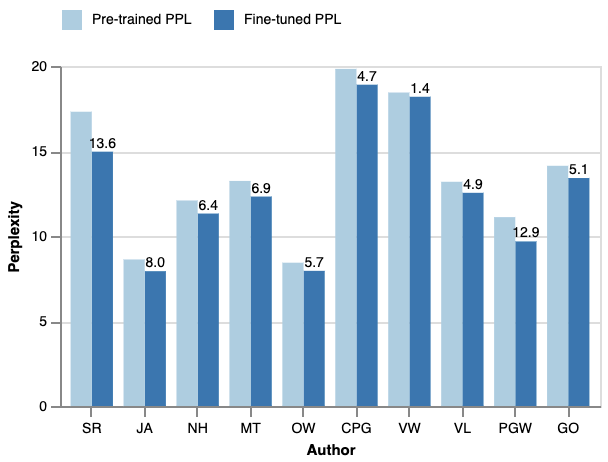}
\caption{Perplexity (PPL) comparison between pre-trained and fine-tuned models across different authors. The number on top of each set of bars indicates the reduction percentage in PPL after fine-tuning. Finetuned models achieve lower scores across all authors.}
\label{fig:ppl}
\end{figure}

PPL of the pre-trained and the corresponding finetuned model are depicted in Figure~\ref{fig:ppl}. The finetuned models consistently exhibit lower perplexity on the validation sets for each author than the base LLaMA-2-7b.
Across all authors, we see an average PPL reduction of 7.0\%.
We see the greatest improvement (13.6\%) for SR---as an 18th-century writer, his language differs the most from the modern English LLaMA was trained on.

\paragraph{Style-embedding Alignment}

\begin{figure}[t]
\centering
\includegraphics[width=0.33\textwidth]{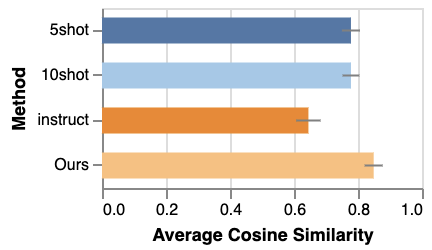}
\caption{Average cosine similarity of baselines and our method between generations and average embeddings across all authors. \textit{StyleTunedLM} archives the highest average similarity.}
\label{fig:cosine}
\end{figure}

Figure~\ref{fig:cosine} illustrates the average cosine similarity between the generated text and author embeddings, with our method achieving the highest average similarities across all authors. In contrast, \textit{instruct} exhibits inconsistencies and difficulties with complex styles. Detailed author-specific performance and confusion matrices from our classifier are available in~\ref{sec:all_style}. We also show the classifier accuracy of each method in Table~\ref{tab:avg_linguistic}.
These results underscore our method's enhanced capability to capture and differentiate authors' writing styles accurately.



\paragraph{Linguistic Alignment}


\begin{table}[!htbp]
\centering
\small
\setlength{\tabcolsep}{5pt}
\begin{tabular}{c|cccc}
\toprule
 &  \textbf{Lexical} & \textbf{Syntactic} & \textbf{Surface} & \textbf{Classifier} \\
\textbf{Method} &  \textbf{\scriptsize (MSE)\(\downarrow\)} & \textbf{\scriptsize (JSD)\(\downarrow\)} & \textbf{\scriptsize (MSE)\(\downarrow\)} & \textbf{Accuracy} \\ \midrule

5shot & 3.80 & 0.07 & 5.43 & 0.693\\ 
10shot & 3.31 & \textbf{0.06} & 4.68 & 0.680 \\ 
instruct & 2.67 & 0.15 & 3.78 & 0.263 \\ 
\textbf{(ours)} & \textbf{1.39} & \textbf{0.06} & \textbf{2.04} & \textbf{0.879}\\ 
\bottomrule
\end{tabular}
\caption{Average linguistic alignment and the BERT classifier accuracy for baselines and our method. \textit{StyleTunedLM} achieves the best overall performance with the lowest errors and highest accuracy.}
\label{tab:avg_linguistic}
\end{table}

Table~\ref{tab:avg_linguistic} presents the average linguistic alignment for our method compared to baselines.
Our approach consistently outperforms the baselines in aligning linguistic features, demonstrating its effectiveness and robustness. Detailed results in~\ref{sec:all_linguistic} reveal our method's proficiency in capturing nuanced writing styles, as it achieves notably low syntactic (0.110) and surface (2.273) errors for VL. This indicates its exceptional ability to replicate the specific word choices and vocabulary patterns of VL's prose.
In contrast, the  \textit{5shot} and \textit{10shot} baselines encounter difficulties with complex styles, with surface errors reaching 25.155 and 22.993 for VL, respectively. This highlights our method's superior capability in replicating intricate stylistic features.
An additional qualitative analysis is available in~\ref{sec:sample}.


\paragraph{Training Size Effects}

Inspired by~\citet{Eder2015DoesSM}, who suggest a minimal size of 5,000 to 10,000 words for stable authorship attribution, we investigate the impact of varying training data sizes---100\%, 70\%, 35\%, and 5\% of 80k tokens, training for three epochs. This simulates scenarios where users have only limited prior work.
Table~\ref{tab:train_size} demonstrates how dataset size affects the model's ability to capture writing styles on average. Training with just 5\% or 35\% of the data leads to significantly low cosine similarity and accuracy, signaling inadequate style learning. As the data size increases, performance is enhanced, evidenced by reduced linguistic errors. 
These findings confirm the relationship between data volume and the model’s capability to learn an author's style.



\begin{table}[t]
\centering
\small
\setlength{\tabcolsep}{2.1pt}
\begin{tabular}{c|cccccc}
\toprule
\makecell{\textbf{\% to full} \\ \textbf{dataset}} &  \textbf{PPL\(\downarrow\)} & \makecell{\textbf{Cosine} \\ \textbf{Sim.}} & \textbf{Acc.} & \makecell{\textbf{Lexical} \\ \textbf{\scriptsize (MSE)\(\downarrow\)}} & \makecell{\textbf{Syntactic} \\ \textbf{\scriptsize (JSD)\(\downarrow\)}} & \makecell{\textbf{Surface} \\ \textbf{\scriptsize (MSE)\(\downarrow\)}} \\ \midrule

5 & 13.47 & 0.57 & 0.11 & 6.04 & 0.12 & 10.02 \\ 
35 & 12.68 & 0.74 & 0.44 & 3.49 & 0.08 & 5.59 \\ 
70 & 12.65 & 0.92 & 0.81 & 1.44 & 0.08 & 2.27 \\ 
100 \textit{(full)} & 12.72 & 0.95 & 0.88 & 1.39 & 0.07 & 2.04 \\ 
\bottomrule
\end{tabular}
\caption{Model performance with different training sizes on average across all authors. Cosine Sim. stands for cosine similarity, and Acc. means classifier accuracy. Performance improves with higher data volume.}
\label{tab:train_size}
\end{table}

\paragraph{Masking out Named Entities}


We craft 50 prompts designed to induce the model to output names, then calculate the total number of names produced and their prevalence in the training data. The prompts are all in the format of ``some words [verb] [name]'' where we delete the names. One example will be ``I don't believe this, said John'' where we delete the name ``John''. We also evaluate whether masking influences the model's style-following ability on corresponding generations. We present the results of PGW and JA in Table~\ref{tab:masking}, focusing on these two authors because ``Jeeves'' is a prevailing character in PGW's work, and similarly, ``Anne'' is a central figure in JA's narratives. Examples and a complete analysis of all authors are available in~\ref{sec:masking_example}. Masking named entities during training has a minimal impact on style learning, with both masked and unmasked models performing similarly. However, the masked model shows lower linguistic errors, implying enhanced generalization. This improvement suggests that masking encourages the model to focus on broader contextual patterns instead of memorizing specific names, effectively reducing overfitting to particular named entities in practical applications. It's worth noting that the effectiveness of masking heavily depends on the accuracy of identifying the targeted named entities.


\begin{table}[t]
\centering
\small
\begin{tabular}{@{}c|cccc@{}}
\toprule
\makecell{\textbf{Ratio} \\ \textbf{\scriptsize (VW:LIMA)}} & \makecell{\textbf{Cosine} \\ \textbf{Similarity}} & \makecell{\textbf{Lexical} \\ \textbf{\scriptsize (MSE)\(\downarrow\)}} & \makecell{\textbf{Syntactic} \\ \textbf{\scriptsize (JSD)\(\downarrow\)}} & \makecell{\textbf{Surface} \\ \textbf{\scriptsize (MSE)\(\downarrow\)}} \\
\midrule
0:1 & 0.57 & 3.45 & 0.11 & 4.74 \\
0.8:1 & 0.59 & 3.42 & 0.10 & 4.32 \\
0.9:1 & 0.64 & 2.17 & 0.07 & 2.86 \\
1:1 & 0.70 & 3.37 & 0.06 & 2.49 \\
\bottomrule
\end{tabular}
\caption{Style alignment for different merging ratios of VW (Virginia Woolf) to LIMA. As the proportion of the style-following adapter increases, performance improves.}
\label{tab:merging_ratios_style}
\end{table}

\paragraph{Merging LoRA Modules}


Merging our \textit{StyleTunedLM} with an adapter tuned on instruction dataset generally not only enables the instruction-following ability but also maintains overall performance across various benchmarks, as detailed in~\ref{sec:merging_result}. We further evaluate the merged model using 20 creative writing prompts collected from three datasets~\cite{huggingface2023instruction, Zhou2023LIMALI, DatabricksBlog2023DollyV2} and quantify its style-following ability. Results in Table~\ref{tab:merging_ratios_style} indicate that higher proportions of the style-following adapter enhance style alignment, reduce linguistic errors, and sustain high cosine similarity. These findings suggest that increasing the style-following adapter's proportion effectively enhances stylistic feature generalization without adversely affecting instruction-following performance.

\section{Conclusion}

In this work, we introduce \textit{StyleTunedLM}, a novel approach leveraging parameter-efficient finetuning (PEFT), aiming to tailor large language models to individual users' stylistic preferences without extensive computational resources. Our results demonstrate that \textit{StyleTunedLM} effectively aligns model outputs with specific stylistic features of different authors, offering significant improvements over traditional methods such as few-shot learning and prompt engineering. We also explore the impact of training data size, content control with masking, and enabling instruction-following capability by merging LoRA modules. 

Future work should conduct additional analysis with writings confirmed to be outside the pre-training corpus to test the generalizability and adaptation capabilities. Furthermore, as we enhance the integration of style and instruction-following modules, developing more refined methods to balance and specify the influence of each component will be crucial for optimizing performance and utility.

\section*{Limitations}
This study primarily focuses on authors whose works are mostly well-represented in the pre-training dataset. We acknowledge the limitation of the generalizability of our findings. The method we proposed has demonstrated robust performance in learning the stylistic nuances of these authors. However, the effectiveness might not extend as effectively to low-resource settings, where the available training data is significantly less. For instance, the model's ability to capture the unique stylistic elements of a user's original work, such as a short essay, remains uncertain. Further work should investigate evaluating style alignment with more user data from diverse and underrepresented authors. 

\section*{Ethics Statement}
While our method is effective in capturing and replicating stylistic nuances, it has raised important ethical concerns. It can be misused to impersonate others, leading to privacy breaches and unauthorized identity use. Additionally, it could be employed to customize models for harmful purposes, such as generating scams or fake news, which could spread misinformation and cause social harm. To prevent misuse, it is crucial to implement strict guidelines and verification processes. By addressing these ethical issues, we aim to ensure our method is used responsibly and beneficially.

\paragraph{\textit{Supplementary Materials Availability Statement}}
\begin{itemize}
    \item 
We will make available for download all author datasets, which include train, validation, and test splits of books chunked into 256 tokens.
    \item
We will release the code needed to re-run the LoRA finetuning for each author.
    \item
We will release the finetuned LoRA weight modules for all experiments as well as the finetuned BERT and Sentence-BERT models used for evaluation.
    \item 
We will release instructions for loading all the above checkpoints into HuggingFace for inference.
\end{itemize}

\bibliography{anthology,custom}

\begin{thebibliography}{33}
\expandafter\ifx\csname natexlab\endcsname\relax\def\natexlab#1{#1}\fi

\bibitem[{Ba{\v{s}}i{\'c} et~al.(2023)Ba{\v{s}}i{\'c}, Banovac, Kru{\v{z}}i{\'c}, and Jerkovi{\'c}}]{bavsic2023chatgpt}
{\v{Z}}eljana Ba{\v{s}}i{\'c}, Ana Banovac, Ivana Kru{\v{z}}i{\'c}, and Ivan Jerkovi{\'c}. 2023.
\newblock Chatgpt-3.5 as writing assistance in students’ essays.
\newblock \emph{Humanities and social sciences communications}, 10(1):1--5.

\bibitem[{Bedington et~al.(2024)Bedington, Halcomb, McKee, Sargent, and Smith}]{bedington2024writing}
Andelyn Bedington, Emma~F Halcomb, Heidi~A McKee, Thomas Sargent, and Adler Smith. 2024.
\newblock Writing with generative ai and human-machine teaming: Insights and recommendations from faculty and students.
\newblock \emph{Computers and Composition}, 71:102833.

\bibitem[{Brown et~al.(2020)Brown, Mann, Ryder, Subbiah, Kaplan, Dhariwal, Neelakantan, Shyam, Sastry, Askell, Agarwal, Herbert-Voss, Krueger, Henighan, Child, Ramesh, Ziegler, Wu, Winter, Hesse, Chen, Sigler, Litwin, Gray, Chess, Clark, Berner, McCandlish, Radford, Sutskever, and Amodei}]{NEURIPS2020_1457c0d6}
Tom Brown, Benjamin Mann, Nick Ryder, Melanie Subbiah, Jared~D Kaplan, Prafulla Dhariwal, Arvind Neelakantan, Pranav Shyam, Girish Sastry, Amanda Askell, Sandhini Agarwal, Ariel Herbert-Voss, Gretchen Krueger, Tom Henighan, Rewon Child, Aditya Ramesh, Daniel Ziegler, Jeffrey Wu, Clemens Winter, Chris Hesse, Mark Chen, Eric Sigler, Mateusz Litwin, Scott Gray, Benjamin Chess, Jack Clark, Christopher Berner, Sam McCandlish, Alec Radford, Ilya Sutskever, and Dario Amodei. 2020.
\newblock \href {https://proceedings.neurips.cc/paper_files/paper/2020/file/1457c0d6bfcb4967418bfb8ac142f64a-Paper.pdf} {Language models are few-shot learners}.
\newblock In \emph{Advances in Neural Information Processing Systems}, volume~33, pages 1877--1901. Curran Associates, Inc.

\bibitem[{Brysbaert et~al.(2014)Brysbaert, Warriner, and Kuperman}]{Brysbaert2014ConcretenessRF}
Marc Brysbaert, Amy~Beth Warriner, and Victor Kuperman. 2014.
\newblock \href {https://api.semanticscholar.org/CorpusID:34985620} {Concreteness ratings for 40 thousand generally known english word lemmas}.
\newblock \emph{Behavior Research Methods}, 46:904--911.

\bibitem[{Clark et~al.(2018)Clark, Cowhey, Etzioni, Khot, Sabharwal, Schoenick, and Tafjord}]{clark2018think}
Peter Clark, Isaac Cowhey, Oren Etzioni, Tushar Khot, Ashish Sabharwal, Carissa Schoenick, and Oyvind Tafjord. 2018.
\newblock Think you have solved question answering? try arc, the ai2 reasoning challenge.
\newblock \emph{arXiv preprint arXiv:1803.05457}.

\bibitem[{Conover et~al.(2023)Conover, Hayes, Mathur, Xie, Wan, Shah, Ghodsi, Wendell, Zaharia, and Xin}]{DatabricksBlog2023DollyV2}
Mike Conover, Matt Hayes, Ankit Mathur, Jianwei Xie, Jun Wan, Sam Shah, Ali Ghodsi, Patrick Wendell, Matei Zaharia, and Reynold Xin. 2023.
\newblock \href {https://www.databricks.com/blog/2023/04/12/dolly-first-open-commercially-viable-instruction-tuned-llm} {Free dolly: Introducing the world's first truly open instruction-tuned llm}.
\newblock Technical report, Databricks.

\bibitem[{Eder(2015)}]{Eder2015DoesSM}
Maciej Eder. 2015.
\newblock \href {https://api.semanticscholar.org/CorpusID:18967587} {Does size matter? authorship attribution, small samples, big problem}.
\newblock In \emph{Digital Scholarship in the Humanities}.

\bibitem[{Face(2023)}]{huggingface2023instruction}
Hugging Face. 2023.
\newblock Huggingfaceh4/instruction-dataset.
\newblock \url{https://huggingface.co/datasets/HuggingFaceH4/instruction-dataset}.

\bibitem[{Feng et~al.(2012)Feng, Banerjee, and Choi}]{feng2012characterizing}
Song Feng, Ritwik Banerjee, and Yejin Choi. 2012.
\newblock Characterizing stylistic elements in syntactic structure.
\newblock In \emph{Proceedings of the 2012 joint conference on empirical methods in natural language processing and computational natural language learning}, pages 1522--1533.

\bibitem[{Gao et~al.(2023)Gao, Tow, Abbasi, Biderman, Black, DiPofi, Foster, Golding, Hsu, Le~Noac'h, Li, McDonell, Muennighoff, Ociepa, Phang, Reynolds, Schoelkopf, Skowron, Sutawika, Tang, Thite, Wang, Wang, and Zou}]{eval-harness}
Leo Gao, Jonathan Tow, Baber Abbasi, Stella Biderman, Sid Black, Anthony DiPofi, Charles Foster, Laurence Golding, Jeffrey Hsu, Alain Le~Noac'h, Haonan Li, Kyle McDonell, Niklas Muennighoff, Chris Ociepa, Jason Phang, Laria Reynolds, Hailey Schoelkopf, Aviya Skowron, Lintang Sutawika, Eric Tang, Anish Thite, Ben Wang, Kevin Wang, and Andy Zou. 2023.
\newblock \href {https://doi.org/10.5281/zenodo.10256836} {A framework for few-shot language model evaluation}.

\bibitem[{Gerlach and Font-Clos(2018)}]{Gerlach2018ASP}
Martin Gerlach and Francesc Font-Clos. 2018.
\newblock \href {https://api.semanticscholar.org/CorpusID:56475903} {A standardized project gutenberg corpus for statistical analysis of natural language and quantitative linguistics}.
\newblock \emph{Entropy}, 22.

\bibitem[{Hasanein and Sobaih(2023)}]{hasanein2023drivers}
Ahmed~M Hasanein and Abu Elnasr~E Sobaih. 2023.
\newblock Drivers and consequences of chatgpt use in higher education: Key stakeholder perspectives.
\newblock \emph{European Journal of Investigation in Health, Psychology and Education}, 13(11):2599--2614.

\bibitem[{Hendrycks et~al.(2020)Hendrycks, Burns, Basart, Zou, Mazeika, Song, and Steinhardt}]{hendrycks2020measuring}
Dan Hendrycks, Collin Burns, Steven Basart, Andy Zou, Mantas Mazeika, Dawn Song, and Jacob Steinhardt. 2020.
\newblock Measuring massive multitask language understanding.
\newblock \emph{arXiv preprint arXiv:2009.03300}.

\bibitem[{Hu et~al.(2021)Hu, Shen, Wallis, Allen-Zhu, Li, Wang, Wang, and Chen}]{hu2021lora}
Edward~J Hu, Yelong Shen, Phillip Wallis, Zeyuan Allen-Zhu, Yuanzhi Li, Shean Wang, Lu~Wang, and Weizhu Chen. 2021.
\newblock Lora: Low-rank adaptation of large language models.
\newblock \emph{arXiv preprint arXiv:2106.09685}.

\bibitem[{Ippolito et~al.(2022)Ippolito, Yuan, Coenen, and Burnam}]{ippolito2022creative}
Daphne Ippolito, Ann Yuan, Andy Coenen, and Sehmon Burnam. 2022.
\newblock Creative writing with an ai-powered writing assistant: Perspectives from professional writers.
\newblock \emph{arXiv preprint arXiv:2211.05030}.

\bibitem[{Lin et~al.(2021)Lin, Hilton, and Evans}]{lin2021truthfulqa}
Stephanie Lin, Jacob Hilton, and Owain Evans. 2021.
\newblock Truthfulqa: Measuring how models mimic human falsehoods.
\newblock \emph{arXiv preprint arXiv:2109.07958}.

\bibitem[{Min et~al.(2022)Min, Lyu, Holtzman, Artetxe, Lewis, Hajishirzi, and Zettlemoyer}]{min2022rethinking}
Sewon Min, Xinxi Lyu, Ari Holtzman, Mikel Artetxe, Mike Lewis, Hannaneh Hajishirzi, and Luke Zettlemoyer. 2022.
\newblock Rethinking the role of demonstrations: What makes in-context learning work?
\newblock \emph{arXiv preprint arXiv:2202.12837}.

\bibitem[{Mirowski et~al.(2023)Mirowski, Mathewson, Pittman, and Evans}]{mirowski2023co}
Piotr Mirowski, Kory~W Mathewson, Jaylen Pittman, and Richard Evans. 2023.
\newblock Co-writing screenplays and theatre scripts with language models: Evaluation by industry professionals.
\newblock In \emph{Proceedings of the 2023 CHI Conference on Human Factors in Computing Systems}, pages 1--34.

\bibitem[{Montani et~al.(2022)Montani, Honnibal, Honnibal, Van~Landeghem, Boyd, Peters, McCann, Samsonov, Geovedi, O'Regan, Altinok, Orosz, Kristiansen, {, Roman}, {Explosion Bot}, {Lj Miranda}, Fiedler, De~Kok, Howard, {, Edward}, {Wannaphong Phatthiyaphaibun}, Tamura, Bozek, {, Murat}, Amery, {Ryn Daniels}, B\"{o}ing, Tippa, and Baumgartner}]{spacy}
Ines Montani, Matthew Honnibal, Matthew Honnibal, Sofie Van~Landeghem, Adriane Boyd, Henning Peters, Paul~O'Leary McCann, Maxim Samsonov, Jim Geovedi, Jim O'Regan, Duygu Altinok, Gy\"{o}rgy Orosz, Søren~Lind Kristiansen, {, Roman}, {Explosion Bot}, {Lj Miranda}, Leander Fiedler, Daniël De~Kok, Grégory Howard, {, Edward}, {Wannaphong Phatthiyaphaibun}, Yohei Tamura, Sam Bozek, {, Murat}, Mark Amery, {Ryn Daniels}, Bj\"{o}rn B\"{o}ing, Pradeep~Kumar Tippa, and Peter Baumgartner. 2022.
\newblock \href {https://doi.org/10.5281/ZENODO.1212303} {explosion/spacy: v3.1.6: Workaround for click/typer issues}.

\bibitem[{Ray(2023)}]{ray2023chatgpt}
Partha~Pratim Ray. 2023.
\newblock Chatgpt: A comprehensive review on background, applications, key challenges, bias, ethics, limitations and future scope.
\newblock \emph{Internet of Things and Cyber-Physical Systems}.

\bibitem[{Reimers and Gurevych(2019)}]{reimers-2019-sentence-bert}
Nils Reimers and Iryna Gurevych. 2019.
\newblock \href {https://arxiv.org/abs/1908.10084} {Sentence-bert: Sentence embeddings using siamese bert-networks}.
\newblock In \emph{Proceedings of the 2019 Conference on Empirical Methods in Natural Language Processing}. Association for Computational Linguistics.

\bibitem[{Sakaguchi et~al.(2021)Sakaguchi, Bras, Bhagavatula, and Choi}]{sakaguchi2021winogrande}
Keisuke Sakaguchi, Ronan~Le Bras, Chandra Bhagavatula, and Yejin Choi. 2021.
\newblock Winogrande: An adversarial winograd schema challenge at scale.
\newblock \emph{Communications of the ACM}, 64(9):99--106.

\bibitem[{Samuel(2023)}]{samual2023what}
Sigal Samuel. 2023.
\newblock \href {https://www.vox.com/future-perfect/23674696/chatgpt-ai-creativity-originality-homogenization} {What happens when chatgpt starts to feed on its own writing?}
\newblock \emph{Vox}.

\bibitem[{Sawicki et~al.(2022)Sawicki, Grzes, Jordanous, Brown, and Peeperkorn}]{sawicki2022training}
Piotr Sawicki, Marek Grzes, Anna Jordanous, Dan Brown, and Max Peeperkorn. 2022.
\newblock Training gpt-2 to represent two romantic-era authors: Challenges, evaluations and pitfalls.
\newblock In \emph{International Conference on Computational Creativity}. Association for Computational Creativity (ACC).

\bibitem[{Syed et~al.(2019)Syed, Verma, Srinivasan, Natarajan, and Varma}]{Syed2019AdaptingLM}
Bakhtiyar Syed, Gaurav Verma, Balaji~Vasan Srinivasan, Anandhavelu Natarajan, and Vasudeva Varma. 2019.
\newblock \href {https://api.semanticscholar.org/CorpusID:202719307} {Adapting language models for non-parallel author-stylized rewriting}.
\newblock \emph{ArXiv}, abs/1909.09962.

\bibitem[{Touvron et~al.(2023)Touvron, Martin, Stone, Albert, Almahairi, Babaei, Bashlykov, Batra, Bhargava, Bhosale, Bikel, Blecher, Ferrer, Chen, Cucurull, Esiobu, Fernandes, Fu, Fu, Fuller, Gao, Goswami, Goyal, Hartshorn, Hosseini, Hou, Inan, Kardas, Kerkez, Khabsa, Kloumann, Korenev, Koura, Lachaux, Lavril, Lee, Liskovich, Lu, Mao, Martinet, Mihaylov, Mishra, Molybog, Nie, Poulton, Reizenstein, Rungta, Saladi, Schelten, Silva, Smith, Subramanian, Tan, Tang, Taylor, Williams, Kuan, Xu, Yan, Zarov, Zhang, Fan, Kambadur, Narang, Rodriguez, Stojnic, Edunov, and Scialom}]{Touvron2023Llama2O}
Hugo Touvron, Louis Martin, Kevin~R. Stone, Peter Albert, Amjad Almahairi, Yasmine Babaei, Nikolay Bashlykov, Soumya Batra, Prajjwal Bhargava, Shruti Bhosale, Daniel~M. Bikel, Lukas Blecher, Cristian~Cant{\'o}n Ferrer, Moya Chen, Guillem Cucurull, David Esiobu, Jude Fernandes, Jeremy Fu, Wenyin Fu, Brian Fuller, Cynthia Gao, Vedanuj Goswami, Naman Goyal, Anthony~S. Hartshorn, Saghar Hosseini, Rui Hou, Hakan Inan, Marcin Kardas, Viktor Kerkez, Madian Khabsa, Isabel~M. Kloumann, A.~V. Korenev, Punit~Singh Koura, Marie-Anne Lachaux, Thibaut Lavril, Jenya Lee, Diana Liskovich, Yinghai Lu, Yuning Mao, Xavier Martinet, Todor Mihaylov, Pushkar Mishra, Igor Molybog, Yixin Nie, Andrew Poulton, Jeremy Reizenstein, Rashi Rungta, Kalyan Saladi, Alan Schelten, Ruan Silva, Eric~Michael Smith, R.~Subramanian, Xia Tan, Binh Tang, Ross Taylor, Adina Williams, Jian~Xiang Kuan, Puxin Xu, Zhengxu Yan, Iliyan Zarov, Yuchen Zhang, Angela Fan, Melanie Kambadur, Sharan Narang, Aurelien Rodriguez, Robert Stojnic, Sergey Edunov, and
  Thomas Scialom. 2023.
\newblock \href {https://api.semanticscholar.org/CorpusID:259950998} {Llama 2: Open foundation and fine-tuned chat models}.
\newblock \emph{ArXiv}, abs/2307.09288.

\bibitem[{Tyo et~al.(2021)Tyo, Dhingra, and Lipton}]{Tyo2021SiameseBF}
Jacob Tyo, Bhuwan Dhingra, and Zachary~Chase Lipton. 2021.
\newblock \href {https://api.semanticscholar.org/CorpusID:237299000} {Siamese bert for authorship verification}.
\newblock In \emph{Conference and Labs of the Evaluation Forum}.

\bibitem[{van Stegeren and My{\'s}liwiec(2021)}]{van2021fine}
Judith van Stegeren and Jakub My{\'s}liwiec. 2021.
\newblock Fine-tuning gpt-2 on annotated rpg quests for npc dialogue generation.
\newblock In \emph{Proceedings of the 16th International Conference on the Foundations of Digital Games}, pages 1--8.

\bibitem[{Verma and Srinivasan(2019)}]{Verma2019ALS}
Gaurav Verma and Balaji~Vasan Srinivasan. 2019.
\newblock \href {https://api.semanticscholar.org/CorpusID:202660671} {A lexical, syntactic, and semantic perspective for understanding style in text}.
\newblock \emph{ArXiv}, abs/1909.08349.

\bibitem[{Wegmann et~al.(2022)Wegmann, Schraagen, and Nguyen}]{Wegmann2022SameAO}
Anna Wegmann, Marijn Schraagen, and Dong Nguyen. 2022.
\newblock \href {https://api.semanticscholar.org/CorpusID:248085045} {Same author or just same topic? towards content-independent style representations}.
\newblock In \emph{Workshop on Representation Learning for NLP}.

\bibitem[{Wei et~al.(2021)Wei, Bosma, Zhao, Guu, Yu, Lester, Du, Dai, and Le}]{Wei2021FinetunedLM}
Jason Wei, Maarten Bosma, Vincent Zhao, Kelvin Guu, Adams~Wei Yu, Brian Lester, Nan Du, Andrew~M. Dai, and Quoc~V. Le. 2021.
\newblock \href {https://api.semanticscholar.org/CorpusID:237416585} {Finetuned language models are zero-shot learners}.
\newblock \emph{ArXiv}, abs/2109.01652.

\bibitem[{Zellers et~al.(2019)Zellers, Holtzman, Bisk, Farhadi, and Choi}]{zellers2019hellaswag}
Rowan Zellers, Ari Holtzman, Yonatan Bisk, Ali Farhadi, and Yejin Choi. 2019.
\newblock Hellaswag: Can a machine really finish your sentence?
\newblock \emph{arXiv preprint arXiv:1905.07830}.

\bibitem[{Zhou et~al.(2023)Zhou, Liu, Xu, Iyer, Sun, Mao, Ma, Efrat, Yu, Yu, Zhang, Ghosh, Lewis, Zettlemoyer, and Levy}]{Zhou2023LIMALI}
Chunting Zhou, Pengfei Liu, Puxin Xu, Srini Iyer, Jiao Sun, Yuning Mao, Xuezhe Ma, Avia Efrat, Ping Yu, L.~Yu, Susan Zhang, Gargi Ghosh, Mike Lewis, Luke Zettlemoyer, and Omer Levy. 2023.
\newblock \href {https://api.semanticscholar.org/CorpusID:258822910} {Lima: Less is more for alignment}.
\newblock \emph{ArXiv}, abs/2305.11206.

\end{thebibliography}
\bibliographystyle{acl_natbib}

\clearpage
\appendix
\onecolumn
\section{Dataset Collection Details}
\subsection{Target Authors}
\label{sec:app-auth}
We shortly introduce each target author including their key literary works, the predominant themes they explore, and their unique contributions to the genres and periods in which they wrote. We use abbreviations in parentheses to represent them throughout this paper.
\begin{itemize}
    \item Samuel Richardson (1689-1761): An English novelist, renowned for pioneering the epistolary form with novels like ``Pamela'' and ``Clarissa''. His works explore the intricate dynamics of personal morality and power within relationships, focusing on domestic virtues and individual dilemmas. \textbf{(SR)}
    \item Jane Austen (1775-1817): An English novelist renowned for her novels like ``Pride and Prejudice'' and ``Emma''. Her works explore the dependence of women on marriage for the pursuit of favorable social standing and economic security. \textbf{(JA)}
    \item Nathaniel Hawthorne (1804-1864): An American novelist and short story writer known for his dark romanticism, notably in ``The Scarlet Letter''. His works often center on the inherent evil and sin of humanity and have moral messages and deep psychological complexity. \textbf{(NH)}
    \item Mark Twain (1835-1910): An American writer, humorist, and essayist famous for ``Adventures of Huckleberry Finn'' and ``The Adventures of Tom Sawyer''. He was praised as the ``greatest humorist the United States has produced''. \textbf{(MT)}
    \item Oscar Wilde (1854-1900): An Irish playwright and novelist, known for his wit and plays like ``The Importance of Being Earnest'' and the novel ``The Picture of Dorian Gray''. \textbf{(OW)}
    \item Charlotte Perkins Gilman (1860-1935): An American feminist, who wrote the short story ``The Yellow Wallpaper'' and other works addressing gendered labor division in society, and the problem of male domination. \textbf{(CPG)}
    \item Virginia Woolf (1882-1941): An English writer and a prominent modernist of the twentieth century known for her novels ``Mrs. Dalloway'' and ``To the Lighthouse''. She pioneered the use of stream of consciousness as a narrative device. \textbf{(VW)}
    \item Vernon Lee (1856-1935): A British writer known for her supernatural fiction and essays on aesthetics such as ``A Phantom Lover''. \textbf{(VL)}
    \item P. G. Wodehouse (1881-1975): An English author best known for his comedic writing, including the Jeeves and Wooster and Blandings Castle series. He was one of the most widely-read humorists of the 20th century. \textbf{(PGW)}
    \item George Orwell (1903-1950): An English novelist and critic best known for ``1984'' and ``Animal Farm.'' His works explore themes of totalitarianism, truth manipulation, and social injustice, significantly shaping modern dystopian literature with his clear, direct prose. \textbf{(GO)}
\end{itemize}

\subsection{Instruction for GPT-4}
\label{sec:gpt4}
We use 100 prompts in total for generation. The first 50 prompts are generated by GPT-4 with the following instruction:

\textit{I want to evaluate 10 models that are finetuned on 10 different authors respectively: Samuel Richardson, Jane Austen, Nathaniel Hawthorne, Mark Twain, Oscar Wilde, Charlotte Perkins Gilman, Virginia Woolf, Vernon Lee, P. G. Wodehouse, George Orwell. First, I have to get some generations from each model. The generations of each model are continuations based on some input prompts, such as the beginning of a sentence. The prompt should not be too long and should be between 6 to 10 words. Based on this experiment design and the characteristics of the 10 authors, please generate 50 prompts for me that are suitable for evaluating all 10 models.}

These 50 prompts are open-ended and versatile, suitable for evaluating models trained on different authors. They encourage diverse narrative responses that reveal each model's ability to capture its author's unique style, themes, and emotional depth. This makes these prompts ideal for our experiments.

\section{Linguistic Alignment Details}
\label{app:linguistic_details}
We evaluate linguistic alignment at three levels: lexical, syntactic, and surface.

Lexical analysis focuses on word-level style choices. In this paper, we consider seven distinct dimensions for lexical analysis: the average numbers of (1) nouns, (2) verbs, (3) adjectives, and (4) unique words per sentence, the average (5) subjectivity scores, and (6) the average number of words with concreteness scores above 3 in a sentence~\citep{Brysbaert2014ConcretenessRF}. This results in a 6-dimensional vector, with each dimension representing one of these features.

Syntactic analysis involves examining the complexity of an author's sentence structures, and determining whether they favor complex or straightforward constructions. We use the algorithm in~\citep{feng2012characterizing} to categorize each sentence into the following five categories: SIMPLE, COMPOUND, COMPLEX, COMPLEX-COMPOUND, and OTHER. This categorization results in a 5-dimensional vector representing the probability distribution over these categories.

Surface analysis focuses on statistical characteristics of the text, such as the average number of (1) commas, (2) semicolons, (3) colons, and the (4) word count per sentence. We also calculate the (5) average length of words. Similar to lexical analysis, it results in a 5-dimensional vector, with each dimension representing one of these features.

\section{Finetuning Details}
\label{app:lora_details}
We conduct our experiments on two A6000 GPUs. Hyperparameters are kept consistent across all methods to ensure a fair comparison, with \verb|learning_rate| set to $5\times10^{-5}$, \verb|num_epoch| set to $3$, \verb|per_gpu_batch_size| set to $4$, and \verb|input_max_token_length| set to $256$.
We ask the model to generate a continuation with $256$ tokens for each input prompt.

\section{Experiment Results}
\subsection{Qualitative Analysis}
\label{sec:sample}

We show some sample generations in the style of Virginia Woolf (VW) and P. G. Wodehouse (PGW) based on the same input prompt in Table~\ref{tab:quality}. The prompt: \textit{I write, my Brother, in the first place, to}, is randomly picked from the evaluation dataset. For better comparison, we color the sentences that show strong alignment with each author's style in their respective colors: \textcolor{myred}{red} for VW and \textcolor{myblue}{blue} for PGW. It highlights the effectiveness of each method in capturing the unique stylistic features of these authors, making it easier to assess the quality of the generated texts visually.

We chose these two authors for our analysis as they show distinct and well-known writing styles, which provides a clear basis for comparison. VW's style is known for its rich, introspective, and figurative language, while PGW's style is characterized by light-hearted, whimsical, and humorous tones. 

From the output, it is evident that \textit{StyleTunedLM} aligns more closely with both authors' styles than the baselines. For VW, it effectively captures the reflective and thoughtful tone, employing vivid imagery and detailed descriptions. An example is the scene of ``a little boy falling off a tricycle'', which evokes strong emotional connections and reflections on human life. In the context of PGW, our method performs equally well, capturing the playful and comedic tone typical of his writing. It uses formal yet amusing language, as seen in phrases like ``I have received the good news of old Fiddler's demise''.

The baselines, however, show varying degrees of success in capturing the target styles. \textit{fewshot} fails to convey the depth of VW's metaphors and introspection and misses the light-hearted tone of PGW. \textit{instruct} presents rich descriptions for VW but can be overly complex, and while it is whimsical for PGW, it is inconsistent in maintaining humor. Overall, \textit{StyleTunedLM} provides a more consistent and accurate representation of both authors' writing styles, making it superior to the other methods.

\begin{table*}[!htbp]
\centering
\setlength{\tabcolsep}{2pt}
{\scriptsize
\begin{tabular}{>{\centering\arraybackslash}m{1.5cm}|m{7cm}|m{7cm}}
\toprule
\centering \textbf{Method} & \textbf{VW} & \textbf{PGW} \\ \midrule
\centering \textbf{5shot} & ...let you know that I am safe. In the second, to thank you. I should never have forgiven myself for remaining away another day, or a minute longer than we had arranged. It is not easy to find people whom you can trust. I am sure of you. You know me, I feel sure. Helen Schumann was sitting at the writing table in her room. She had just finished reading her brother's letter. Her father had gone down to the station where the telegram from her mother and the letter from her brother awaited him. \textcolor{myred}{She remained standing in front of the empty writing table, with her hands clasped behind her head, thinking, and then turned away from the window, from which she could see the trees of a wood against the evening light, and began slowly to pace to and fro across the floor.} She had left home before the summer vacation had begun. Now there was still three months until autumn. But that did not trouble Helen. She knew what she intended to do. She walked up and down, smiling to herself. And in a way, too, she laughed at her father. What had he ever done? What could he do now? He had lost control over her & ...let you know \textcolor{myblue}{how much I enjoy reading your columns in the Gazette. It is a singularly entertaining paper, and I should not be without it for a day.} You can rely on me always to appreciate any point which may be new to me, and I feel sure that you will not grudge the time necessary to explain anything that may require explanation. Secondly, the fact is, dear Brother, that ever since my nephew's visit I have been considering taking up some work for myself, as you urged, so that I may have something to talk about other than the weather and the crops. In my case it will be not the weather nor the crops, but my experiences in journalism, and I assure you that they are worth recording, and worth telling others about. For instance: In my early days, when I was just beginning, it used to be thought very wrong of me to try to sell articles to editors. They would say that it was taking advantage of an editor's poverty, that the editorial staff was quite big enough already, and so on. Now, however, people do not hesitate to send round to editors offering articles, and in this connection I can mention that I have recently sent around to the \\ \midrule
\centering \textbf{10shot} & ...let you know that I am safe. In the second, to thank you. \textcolor{myred}{I want you to forgive this letter--all letters are, after a fashion, selfish;} but I feel as if you wanted to know about my adventures. I am quite safe--amused, interested, sorry for some people; and happy, very happy. That's about it, dear Brother. \textcolor{myred}{But we've talked so often about life that I feel somehow as if it would be strange not to write to you--if only to tell you that life's very good; that it is worth living; and that every time a person writes to another, "Well, and how does life treat you?" it's a compliment. It means a person cares whether another lives happily.} You don't. So you won't be surprised that I've written. But I wanted to tell you that I haven't been so happy since--since you know when. My days pass swiftly, but slowly too. For one thing, I see more than ever. People speak. They don't talk as you say they do. I've learned a lot in five weeks. But then London--London, I mean & ...let you know how I am getting on; secondly, to thank you most sincerely for the check which accompanied your letter of December 18th. \textcolor{myblue}{I must admit that I am making rather slow progress in the world of finance. I am only in the fifth grade in the bank. There are twenty others higher than me and I fancy it may take me a good bit longer to reach the top.} My chief duty consists of handling all correspondence with members and depositors in the matter of their savings accounts and the issue of dividends. \textcolor{myblue}{I find it quite interesting, though I confess that the amount of figures and sums involved sometimes makes my head swim.} I have been reading some articles in the financial pages of the papers concerning the state of British finances and the depression of trade. You don't want me to talk about your wife any more, but I must tell you that she called up my sister-in-law, Mrs Cummings, yesterday to speak of you. Mrs Cummings seems to have been rather annoyed, as I gather she thought you were still angry with her. She says it's hard to explain things on the phone without seeming to intrude \\ \midrule
\centering \textbf{instruct} & ...inform you that I have taken up my pen once more, to continue the tale of our lives. The wind whispers secrets in my ear, and I listen intently, for it is through its gentle caress that I find inspiration. As I write, the world around me fades into the background, and I am transported to realms of thought and imagination. \textcolor{myred}{My mind is a kaleidoscope of colors, shapes, and sounds, each one vying for dominance in the ever-changing tapestry of my creative endeavors.} And so, dear Brother, I must confine myself to the page, lest I lose myself altogether in the labyrinth of my own mind. \textcolor{myred}{For in this world of words, I find solace, comfort, and the occasional glimmer of madness. But such is the price we pay for the luxury of creation, is it not? Yet still, I would not have it any other way, for in the end, it is the act of writing itself that sustains us, that keeps us alive and connected to the universe at large.} And so, with pen poised and mind afire, I embark upon the next chapter of our collective narrative & ...invite you to \textcolor{myblue}{partake of a most delightful repast at my humble abode}. I trust that you will find the viands and beverages presented before you to be nothing short of exquisite, for I have taken the utmost care in their preparation. But tell me, my dear fellow, how have you been faring of late? \textcolor{myblue}{Have you encountered any particularly egregious examples of pomposity or absurdity in your daily interactions? Do share, I am simply bursting with curiosity on the subject.} And speaking of curiosities, have you had the opportunity to peruse the latest installment of the 'Gas-gentleman's Quarterly'? I must say, the piece on the etiquette of tooting one's own horn is quite scintillating. Now, if you will excuse me, I must take my leave. \textcolor{myblue}{A most charming young lady has arrived at the doorstep, claiming to have lost her way. I fear she may require my assistance in navigating the byzantine labyrinth of my mansion's corridors.} Until we meet again, my dear Brother, do enjoy the rest of your \\ \midrule
\centering \textbf{StyleTunedLM} & ...\textcolor{myred}{tell you of a change that has come over me; as the first act on the stage, before one can say what follows, has to be struck.} I was not ill, nor unhappy (thank God), but only, \textcolor{myred}{it seemed so strange to see the world from another point of view, and everything had changed--the trees, the grass--it must be autumn!} Only one can't think about anything at all just yet, can one? But, my dear, I felt this way once, coming out of church after your marriage, when I had seen myself suddenly as other people saw me for the first time, but with less regret than pity then; though I loved you and thought you perfect--as I think you still!--it did not seem so much. But now, here we are. And do you know what happened? \textcolor{myred}{The little boy next door--Peter Walsh--fell off his tricycle into our garden, where he lay howling and covered with blood. There is no one like a child for making you feel life again; and it is odd, isn't it, that you should find life in a crying child?} It made me wonder if I ever could have loved Richard, if I hadn't known it & ...apprise you that \textcolor{myblue}{I have received the good news of old Fiddler's demise from a telegram addressed to me at "Potts \& Sons, Newcastle-upon-Tyne."} It arrived this morning by registered post. You may recall that this was one of the things you were to do for me in your capacity as head clerk in our office. \textcolor{myblue}{In point of fact it would seem that you did not even bother to send off the requisite form until about three days after the event--I received no fewer than four telegrams on the subject before I had been able to get over to London and start my legal proceedings.} However, the thing is done now; and there remains only to take up with old Fiddler's executors what we may. \textcolor{myblue}{And here we strike what appears to be a slight snag.} I don't know if I've mentioned the name of Mr. Bingo Little to you before? No? Well, he's old Fiddler's sole next of kin. He is a fellow who lives out East somewhere, where they like him very much. I met him once or twice while I was staying with old Fiddler in the early stages of his illness \\ \bottomrule
\end{tabular}
}
\caption{Sample generations of baselines and our method in the style of VW and PGW. Sentences that show strong alignment are colored in \textcolor{myred}{red} for VW and \textcolor{myblue}{blue} for PGW.}
\label{tab:quality}
\end{table*}


\subsection{t-SNE Analysis}
\label{sec:all_tsne}
We show the t-SNE visualizations using pairwise loss on the training and test datasets, and generation of \textit{StyleTunedLM}. The distinct clusters in the training dataset (\textbf{left}) suggest that the style attribution model we trained has learned to distinguish between the style embeddings of different authors. Clusters in the testing dataset (\textbf{middle}) and the generation (\textbf{right}) show certain overlap, but they share consistent clustering patterns, suggesting that the model can be effectively used to evaluate embedding similarity.

\begin{figure*}[!htbp]
\centering
\includegraphics[width=1.0\textwidth]{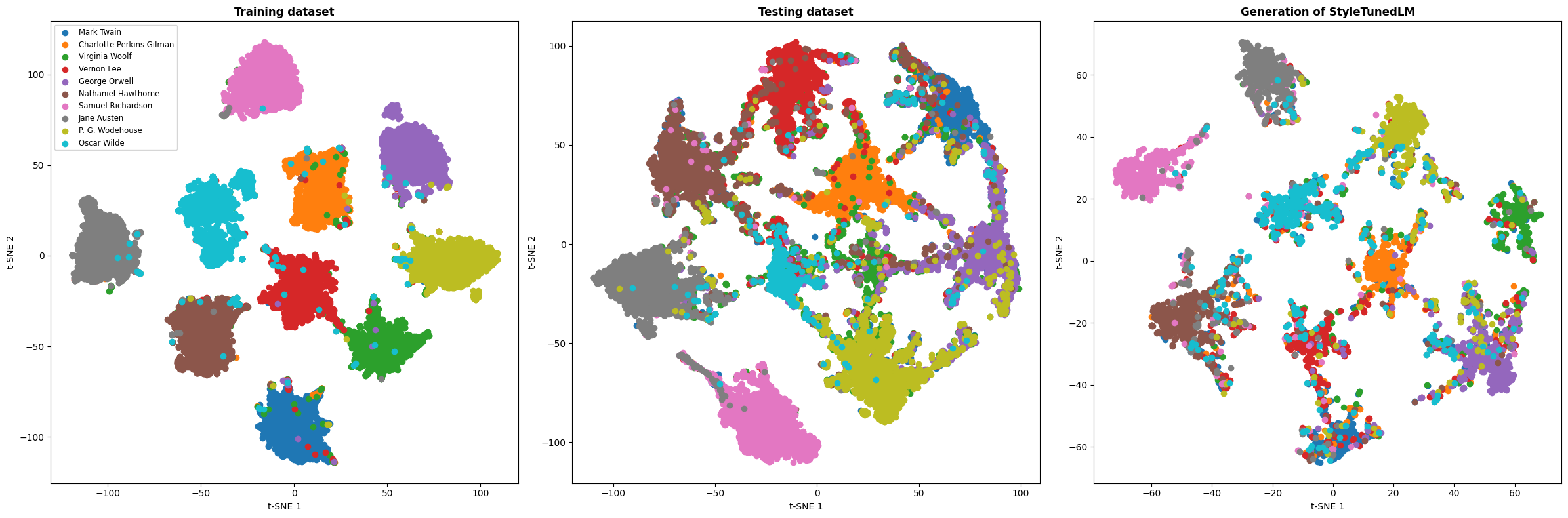}
\caption{t-SNE on training, test, and generation of our method with pairwise loss.}
\label{fig:tsne}
\end{figure*}

\subsection{Style-embedding Alignment Analysis}
\label{sec:all_style}

\begin{figure*}[!htbp]
\centering
\includegraphics[width=1.0\textwidth]{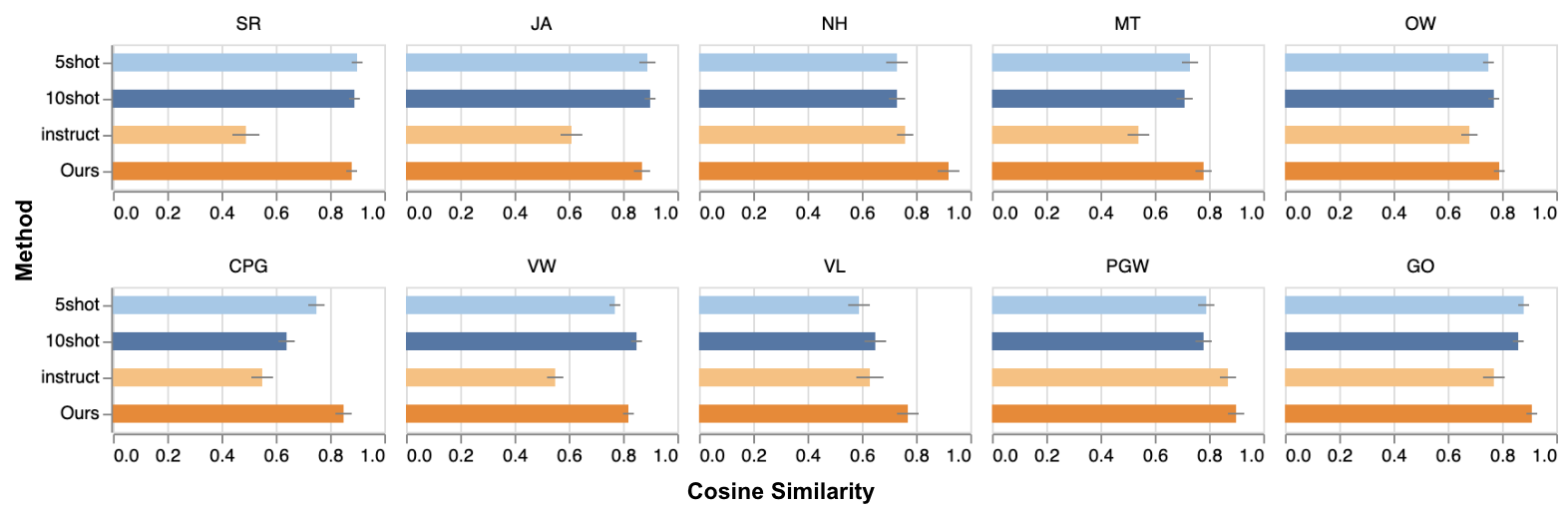}
\caption{Cosine similarity of baselines and our method between generation and the average embedding for each author. \textit{StyleTunedLM} archives the highest similarities with most authors.}
\label{fig:all_cosine}
\end{figure*}

Figure~\ref{fig:all_cosine} illustrates the cosine similarity for each author. Our method consistently achieves the highest scores on most authors, effectively capturing nuanced features such as Nathaniel Hawthorne's (NH) complex symbolism and intricate sentence structures. In contrast, \textit{5shot} and \textit{10shot} show moderate performance, while \textit{instruct} frequently underperforms, particularly in learning complex stylistic elements.

\begin{figure*}[h]
\centering
\includegraphics[width=1.0\textwidth]{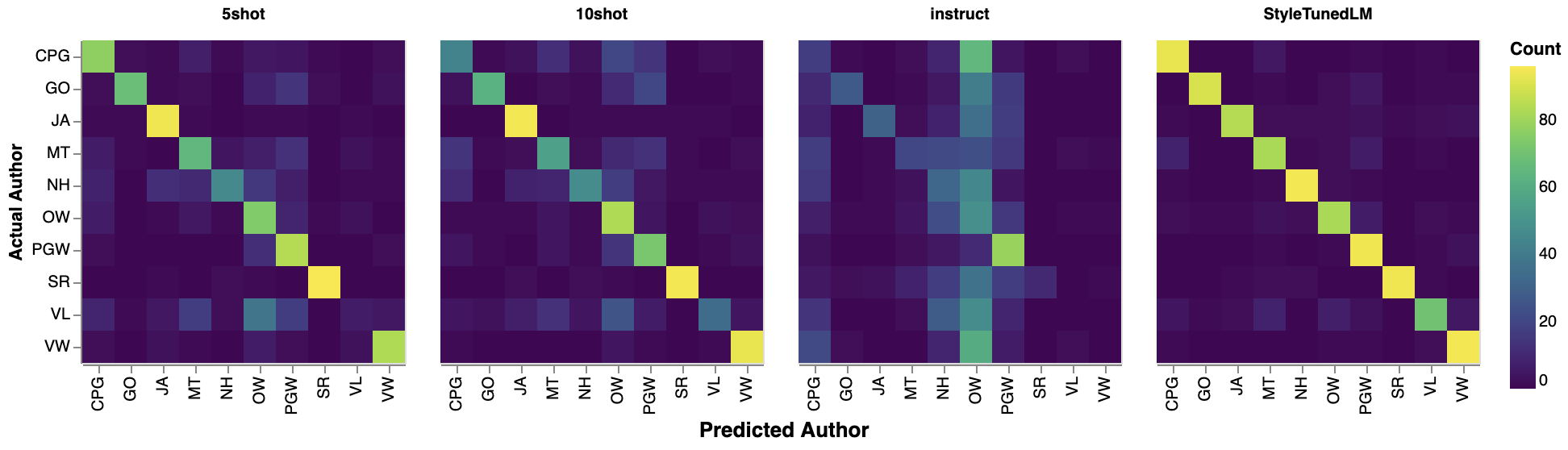}
\caption{Confusion matrices of baselines and our method. \textit{StyleTunedLM} achieves the highest classification accuracies across all authors.}
\label{fig:confusion}
\end{figure*}

Confusion matrices in Figure~\ref{fig:confusion} confirm similar findings, showing that \textit{StyleTunedLM} attains the highest classification accuracy at 87.9\%, significantly outperforming other baselines. 

\subsection{Linguistic Alignment Analysis}
\label{sec:all_linguistic}

We show the detailed linguistic alignment analysis results in Table~\ref{tab:evaluation}. \textit{Lexical} and \textit{Surface} are measured by MSE, while \textit{Syntactic} is measured by JSD as described in \S\ref{sec:framework}. \textit{StyleTunedLM} generally shows superior or competitive performance across the three levels. For instance, it exhibits the lowest syntactic error at 0.010 and a notably reduced surface error at 6.690 for Nathaniel Hawthorne (NH), indicating its effectiveness in capturing the stylistic nuances of the author's writing. Similarly, for Mark Twain (MT), it improves syntactic alignment with the lowest error of 0.047 and significantly reduces the surface error to 1.849. These results suggest that \textit{StyleTunedLM} effectively minimizes deviations from the target author's style.

\begin{table*}[h]
\centering
\small 
\begin{tabular}{ccccc}
\toprule
\textbf{Author} & \textbf{Method} & \textbf{Lexical \scriptsize(MSE)\(\downarrow\)} & \textbf{Syntactic \scriptsize(JSD)\(\downarrow\)} & \textbf{Surface \scriptsize(MSE)\(\downarrow\)} \\ 
 \midrule
\multirow{4}{*}{\textbf{SR}} & 5shot & 0.051 & 0.106 & 0.069 \\ 
 & 10shot & \textbf{0.045} & \textbf{0.039} & \textbf{0.002} \\ 
 & instruct & 0.180 & 0.132 & 0.249 \\ 
 & StyleTunedLM & 0.220 & 0.083 & 0.161 \\ \midrule
\multirow{4}{*}{\textbf{JA}} & 5shot & 8.546 & 0.040 & 11.701 \\ 
 & 10shot & 4.443 & \textbf{0.021} & 6.207 \\ 
 & instruct & 4.710 & 0.127 & 7.469 \\ 
 & StyleTunedLM & \textbf{3.466} & 0.029 & \textbf{5.745} \\ \midrule
\multirow{4}{*}{\textbf{NH}} & 5shot & 2.280 & 0.063 & 2.654 \\ 
 & 10shot & 3.738 & 0.082 & 4.455 \\ 
 & instruct & \textbf{0.185} & 0.109 & \textbf{0.110} \\ 
 & StyleTunedLM & 4.067 & \textbf{0.010} & 6.690 \\ \midrule
\multirow{4}{*}{\textbf{MT}} & 5shot & 6.404 & 0.082 & 8.717 \\ 
 & 10shot & 7.020 & 0.079 & 9.551 \\ 
 & instruct & 1.664 & 0.121 & 3.294 \\ 
 & StyleTunedLM & \textbf{1.132} & \textbf{0.047} & \textbf{1.849} \\ \midrule
\multirow{4}{*}{\textbf{OW}} & 5shot & \textbf{0.180} & \textbf{0.089} & \textbf{0.002} \\ 
 & 10shot & 0.195 & 0.109 & 0.030 \\ 
 & instruct & 9.115 & 0.192 & 11.001 \\ 
 & StyleTunedLM & 0.690 & 0.116 & 0.532 \\ \midrule
\multirow{4}{*}{\textbf{CPG}} & 5shot & 2.534 & \textbf{0.023} & 2.986 \\ 
 & 10shot & 1.321 & 0.039 & \textbf{1.169} \\ 
 & instruct & \textbf{1.231} & 0.181 & 1.741 \\ 
 & StyleTunedLM & 1.789 & 0.042 & 2.456 \\ \midrule
\multirow{4}{*}{\textbf{VW}} & 5shot & 1.740 & 0.051 & 1.846 \\ 
 & 10shot & 1.783 & \textbf{0.044} & 1.843 \\ 
 & instruct & 6.074 & 0.263 & 8.858 \\ 
 & StyleTunedLM & \textbf{0.613} & 0.086 & \textbf{0.324} \\ \midrule
\multirow{4}{*}{\textbf{VL}} & 5shot & 15.255 & 0.107 & 25.155 \\ 
 & 10shot & 13.965 & 0.133 & 22.993 \\ 
 & instruct & \textbf{1.092} & 0.222 & 2.710 \\ 
 & StyleTunedLM & 1.312 & \textbf{0.110} & \textbf{2.273} \\ \midrule
\multirow{4}{*}{\textbf{PGW}} & 5shot & 0.138 & 0.055 & 0.049 \\ 
 & 10shot & \textbf{0.087} & \textbf{0.027} & \textbf{0.002} \\ 
 & instruct & 1.417 & 0.047 & 1.383 \\ 
 & StyleTunedLM & 0.477 & 0.075 & 0.336 \\ \midrule
\multirow{4}{*}{\textbf{GO}} & 5shot & 0.878 & 0.063 & 1.137 \\ 
 & 10shot & 0.523 & 0.063 & 0.565 \\ 
 & instruct & 1.001 & 0.095 & 1.020 \\ 
 & StyleTunedLM & \textbf{0.130} & \textbf{0.041} & \textbf{0.009} \\
 \bottomrule
\end{tabular}
\caption{Lexical, syntactic, and surface errors of baselines and our method for each author. \textit{StyleTunedLM} consistently demonstrates superior performance in minimizing three levels of errors.}
\label{tab:evaluation}
\end{table*}

\subsection{Masking during Training}
\label{sec:masking_example}

We show the complete experiment results of masking on all authors in Table~\ref{tab:masking_all}. We then present two pairs of examples generated by models finetuned on the books of P. G. Wodehouse (PGW) in Table~\ref{tab:masking_output}, with and without the masking technique during training. In these examples, names immediately following the prompts are highlighted in \textbf{bold}, and names that also appear in the training data are marked in \textit{italics}.
Without masking, the model frequently recalls names like ``Bingo'' and ``Aunt Agatha'', which are prevalent in the training data, incorporating them as characters in the generated outputs. Conversely, with masking applied, the model avoids overfitting to specific names in the training data, opting for other names and pronouns in its generation. To be noted, the differences between the content generated with and without masking can be attributed to a high temperature setting (0.9) during generation, which increases creativity and reduces determinism. When a different name is predicted due to masking, the model generates a continuation based on this new context, leading to a noticeable divergence in the narratives.

\begin{table*}[t]
\centering
\small
\begin{tabular}{cc|cc|cccccc}
\toprule
\textbf{Author} & \textbf{Method} & \textbf{\% in training} & \textbf{\# of names} & \textbf{PPL\(\downarrow\)} & \makecell{\textbf{Cosine} \\ \textbf{Sim.}} & \textbf{Acc.} & \makecell{\textbf{Lexical} \\ \textbf{\scriptsize (MSE)\(\downarrow\)}} & \makecell{\textbf{Syntactic} \\ \textbf{\scriptsize (JSD)\(\downarrow\)}} & \makecell{\textbf{Surface} \\ \textbf{\scriptsize (MSE)\(\downarrow\)}} \\ \midrule

\multirow{2}{*}{SR} & w/o masking & 0.58 & 62 & 14.96 & 0.92 & 0.88 & 0.44 & 0.08 & 2.36 \\ 
& w/ masking & 0.41 & 59 & 15.33 & 0.95 & 0.82 & 0.36 & 0.05 & 1.65 \\ 
\multirow{2}{*}{JA} & w/o masking & 0.61 & 62 & 7.93 & 1.0 & 1.0 & 7.72 & 0.04 & 12.53 \\ 
& w/ masking & 0.45 & 85 & 8.02 & 0.90 & 0.76 & 4.62 & 0.03 & 7.49 \\ 
\multirow{2}{*}{NH} & w/o masking & 0.57 & 72 & 11.32 & 1.0 & 1.0 & 5.75 & 0.05 & 9.56 \\ 
& w/ masking & 0.29 & 96 & 11.39 & 0.97 & 0.72 & 5.23 & 0.04 & 8.70 \\
\multirow{2}{*}{MT} & w/o masking & 0.26 & 53 & 12.32 & 0.93 & 0.80 & 4.01 & 0.04 & 7.16 \\
& w/ masking & 0.23 & 44 & 12.71 & 0.93 & 0.76 & 6.45 & 0.03 & 11.01 \\
\multirow{2}{*}{OW} & w/o masking & 0.33 & 104 & 8.05 & 0.94 & 0.88 & 0.35 & 0.12 & 0.01 \\
& w/ masking & 0.19 & 85 & 7.95 & 0.86 & 0.74 & 1.19 & 0.16 & 0.49 \\
\multirow{2}{*}{CPG} & w/o masking & 0.46 & 41 & 18.90 & 1.0 & 1.0 & 3.43 & 0.03 & 5.07 \\
& w/ masking & 0.12 & 99 & 19.25 & 0.96 & 0.84 & 2.39 & 0.03 & 3.84 \\
\multirow{2}{*}{VW} & w/o masking & 0.46 & 84 & 18.19 & 0.97 & 0.98 & 0.49 & 0.12 & 0.20 \\
& w/ masking & 0.27 & 78 & 18.07 & 0.88 & 0.80 & 0.48 & 0.11 & 0.07 \\
\multirow{2}{*}{VL} & w/o masking & 0.20 & 81 & 12.54 & 0.86 & 0.62 & 12.51 & 0.05 & 24.32 \\
& w/ masking & 0.16 & 58 & 12.75 & 0.79 & 0.48 & 13.67 & 0.05 & 26.85 \\
\multirow{2}{*}{PGW} & w/o masking & 0.50 & 68 & 9.68 & 1.0 & 1.0 & 0.18 & 0.07 & 0.01 \\ 
& w/ masking & 0.23 & 91 & 10.46 & 0.98 & 0.90 & 0.16 & 0.07 & 0.11 \\ 
\multirow{2}{*}{GO} & w/o masking & 0.19 & 85 & 13.41 & 1.0 & 0.98 & 0.82 & 0.06 & 1.83 \\
& w/ masking & 0.09 & 91 & 13.25 & 0.76 & 0.16 & 0.72 & 0.06 & 1.69 \\
\bottomrule
\end{tabular}
\caption{Model performance with and without masking during training of all ten authors.}
\label{tab:masking_all}
\end{table*}

\begin{table*}[!htbp]
\centering
\small
\begin{tabular}{p{0.45\textwidth}|p{0.45\textwidth}}
\toprule
\textbf{w/o masking} & \textbf{w/ masking} \\ 
\midrule
"Hush, naughty one!" returned \textbf{Aunt Agatha} firmly. "I don't mind your making fun of the old lady and the beastly pink gown she was wearing, but I must insist that you call her 'madam.'" "Right-o!" said \textit{Jeeves}. "The usual thing? The old girl?" "Well, as a rule the maid brings it." "Oh, very well. What time do they bring it?" It may have been the effect of my having got through so much cold chicken on the previous day, but I suddenly felt in rather an offish mood. "You mean that this--this woman who calls herself \textit{Mrs. Bingo Little} and lives at number ten, Foley Street, is waiting to come down here for my clothes?" There was silence from \textit{Jeeves}. "She mayn't be there, sir," he said. "It is possible that the telegram may not reach her in time; or even if it does, there is no guarantee that she will arrive punctually at half past twelve. She may be later than that." "How later?" I asked sharply. "Well, perhaps as late as three o'clock, sir," said \textit{Jeeves} & "Hush, naughty one!" returned \textbf{the old bird} in a voice so shrill and high-pitched that I had to clap my hands to my ears. "I am a bishop's daughter." "That's right! That's where you went wrong. You see, if you'd said you were a clergyman's daughter--that would have been different. The thing for us to do is to find some clergyman who'll be willing to stand godfather for the kid, eh?" Old Egg shot a glance of loathing at the infant as it continued to make noises like an engine under stress. "You will never get any clergyman to stand godfather," he hissed. "A clergyman will not become entangled in such a degrading business!" "Not a soul!" chimed in young Egg. The child looked up from its gravel and stared at him with mild surprise. Then the father gave it another little slap. "If you try it on," he went on, "you may depend upon it there will be scandal." "Scandal? How can there be scandal when the child has no name?" "When the child grows up the story of this shameful affair will \\ \midrule
"For Heaven's sake, what news do you bring?" screamed \textbf{Bingo}. "Anything about my cousin?" "Yes." I could see \textit{Bingo} was quite shaken by this. "Well, well! Go on!" said he. It was not a pleasant picture of a boy who had been to the bad places, and who looked as though he had been eating too much sugar. But it was all right in its way. There is something about a badly-brought-up youngster that puts life into your day. I suppose it is because we have had so little experience with them; they are a rare commodity in England, where there are few children and most of those have been thoroughly civilised at birth. I am not saying that Eustace Clowes was a joyful addition to society. His main characteristic seemed to be his tendency to slap me on the back when we met in the hall or the library, or wherever we happened to run across each other. He did this several times before I got accustomed to the idea, and every time it jolted me. After dinner one evening, just before we started for the study, he came dashing up and slapped me heartily on the ribs. "Good egg," said \textit{Eustace} & "For Heaven's sake, what news do you bring?" screamed \textbf{the girl}. "The news that I have been a long time getting round to telling you," said I. It wasn't my fault. The fact is, as soon as I got back from town I went straight off to call on Mrs. Beavington in order to hear more about poor old old Mr. and Mrs. \textit{Pringle}; and there, in their sitting-room, were the entire Pringles themselves. The only other person present was the Rev. Cecil St. Bessent, who had arrived at the last moment. He was having tea with them when I arrived. Well, sir, let me tell you that this tea was no ordinary tea. Tea doesn't often affect me strongly one way or another. But this tea affected me deeply--in an emotional sense--and it knocked everything else into a cocked hat. Even after all these years I can still see before me those three persons--Mrs. \textit{Pringle} and Mr. \textit{Pringle} and the Rev. Cecil--with the silver teapot between them. And you know how quickly your memory fades away after a time. Oh, yes. All sorts of things fade away--but not that scene. After that cup of tea \\ \bottomrule
\end{tabular}
\caption{Sample generation with and without masing during training. Names following the prompts are highlighted in \textbf{bold}, and other names that also appear in the training data are marked in \textit{italics}.}
\label{tab:masking_output}
\end{table*}

\subsection{LoRA Module Merging for Enabling Instruction-Following Ability}
\label{sec:merging_result}
Table~\ref{tab:merging_ratios_openllm} shows the performance of models with different merging ratios on several benchmarks~\citep{eval-harness, hendrycks2020measuring, sakaguchi2021winogrande, clark2018think, zellers2019hellaswag, lin2021truthfulqa}. While there are minor fluctuations, the scores remain relatively stable across different merging ratios, indicating that such merging is viable without detrimental effects. 

\begin{table*}[h]
\centering
\small
\begin{tabular}{cccccccc}
\toprule
\makecell{\textbf{Ratio} \\ \textbf{\scriptsize(VW:LIMA)}} & \textbf{MMLU} & \textbf{WinoGrande} & \makecell{\textbf{ARC} \\ \textbf{Easy}} & \makecell{\textbf{ARC} \\ \textbf{Challenge}} & \textbf{HellaSwag} & \makecell{\textbf{TruthfulQA} \\ \textbf{MC1}} & \makecell{\textbf{TruthfulQA} \\ \textbf{MC2}} \\
\midrule
0:1 & 0.336 & 0.639 & 0.710 & 0.457 & 0.550 & 0.246 & 0.373 \\
0.8:1 & 0.339 & 0.640 & 0.706 & 0.462 & 0.546 & 0.239 & 0.363 \\
0.9:1 & 0.340 & 0.646 & 0.705 & 0.466 & 0.545 & 0.236 & 0.362 \\
1:1 & 0.340 & 0.647 & 0.707 & 0.462 & 0.544 & 0.236 & 0.361 \\
\bottomrule
\end{tabular}
\caption{5-shot performance for different merging ratios of VW (Virginia Woolf) to LIMA. By merging, the model is enabled with instruction-following ability and different ratios have no detrimental impact on the performance.}
\label{tab:merging_ratios_openllm}
\end{table*}

\end{document}